# EXTRACTING EMOTION PHRASES FROM TWEETS USING BART

Mahdi Rezapour

July 27, 2024

## ABSTRACT

Sentiment analysis is a natural language processing task that aims to identify and extract the emotional aspects of a text. However, many existing sentiment analysis methods primarily classify the overall polarity of a text, overlooking the specific phrases that convey sentiment. In this paper, we applied an approach to sentiment analysis based on a question answering framework. Our approach leverages the power of Bidirectional Autoregressive Transformer (BART), a pre-trained sequence-to-sequence model, to extract a phrase from a given text that amplifies a given sentiment polarity. We create a natural language question that identifies the specific emotion to extract and then guide BART to pay attention to the relevant emotional cues in the text. We use a classifier within BART to predict the start and end positions of the answer span within the text, which helps to identify the precise boundaries of the extracted emotion phrase. Our approach offers several advantages over most sentiment analysis studies, including capturing the complete context and meaning of the text and extracting precise token spans that highlight the intended sentiment. We achieved an end loss of 87% and Jaccard score of 0.61.

## 0.1 Introduction

Sentiment analysis involves identifying and extracting emotions expressed in natural language texts, such as tweets, reviews, or articles. This task has many applications in various domains, such as social media analysis [15], customer feedback [3], and opinion mining [2]. However, sentiment analysis poses several challenges, such as the complexity and diversity of natural language [16], the subtlety and context-dependence of emotions [11], and the difficulty of capturing the full sentiment conveyed by a text [18].

One common approach to sentiment analysis is to extract emotion-laden phrases from the text using linguistic rules or part-of-speech patterns [7-10]. These phrases are then used to infer the overall sentiment polarity (positive, negative, or neutral) of the text. However, this approach has some limitations, such as missing subtle nuances [22], failing to capture the full context [12], and not accurately conveying the intended sentiment. For example, the phrase "I love this product" may have different meanings depending on the tone, sarcasm, or irony of the speaker.

To address these limitations, we employ an approach to sentiment analysis based on a question-answering framework. Our approach leverages the power of BART (Bidirectional and Auto-Regressive Transformers) [6], a state-of-the-art pre-trained sequence-to-sequence model, to extract a phrase from a given text that amplifies a given sentiment polarity.

BART is a denoising autoencoder that is trained by reconstructing the original text from corrupted text by an arbitrary noising function [17]. BART greatly resembles T5 (Text-To-Test Transfer Transformer) [9]in terms of encode-decode architecture and denoising pre-training objective. BART has two main advantages for our task: first, it can process text bidirectionally, similar to (Bidirectional Encoder Representations from Transformers) BERT [13], allowing it to capture the comprehensive context and meaning of the text; second, it can generate text left-to-right, similar to Generative Pre-trained Transformer (GPT) [4], enabling it to create phrases that directly express the targeted emotion.

The implemented approach consists of two steps: first, we formulate a natural language question that aims at the specific emotion to be extracted. Questions guides BART to focus on the relevant emotional cues within the given text. Second, we feed the combined question and text to the BART encoder, which encodes them into a latent representation, the method discusses that in detail. Then, we use the BART decoder to find a phrase location that answers the question and expresses the targeted emotion. To do so we use a classifier within BART to predict the start and end indices of the answer span within the text, which helps to identify the precise boundaries of the extracted emotion phrase.

## 0.2 Methodology

In this section, we describe the model, the data preprocessing, the loss function, and the answer extraction that we used for our question answering task. We used the BART model [20] as our sequence-to-sequence model. BART is a pre-trained model that uses a bidirectional encoder like BERT [8], and a left-to-right decoder like GPT [1].

### 0.2.1 Data Preprocessing

Before feeding the input to the model, we preprocessed the text data to make it suitable for BART's input and output format. We followed the general steps below:

- We added a space character at the beginning of each selected text in the train dataset as train['selected_text']= " " + train['selected_text']. This helps the tokenizer to split the selected text correctly, especially if it starts with a punctuation mark or a special token.
- We created a new variable called processed_input_train, which is a string that contains the prefix "extract: ", followed by the sentiment of the question, followed by two special tokens "</s><s>", followed by the context of the question. The context is the original text from which the answer span is extracted. The prefix and the special tokens are used to indicate the task and the format of the input for the BART model. For example, processed_input_train = "extract: positive</s><s>context: I love this movie!".
- On the other hand, the response or the extracted span is done like train['selected_text'] = train['selected_text'] + ' </s>'. For instance, for the above sentence, the target might just be "love" so the model is expected to extract that term. When training the model, the target_ids tensor is created by appending the </s> token to the end of each target sequence. This is done to ensure that the model learns to generate sequences that end with the </s> token.

We used the BartTokenizer class, which is inherited from the PreTrainedTokenizer class to tokenize the input text. That tokenizer automatically adds the special tokens "<s>" and "</s>" to the beginning and the end of the input sequence.

To find the start position of the selected text in the input sequence, we looped over the elements of the output sequence, which are the ids of the tokens in the selected text. For each element, we searched for the first matching index in the input sequence and assigned it to a variable called start_position.

We then broke the loop, if we had found the start position of the selected text in the input sequence as we did not need to check the rest of the output sequence. We added 8 to the start_position, to account for the four special tokens that are added by the BartTokenizer, which have a total length of 8. This gave us the correct position of the selected text in the original input sequence.

To find the end position of the selected text in the input sequence, we used a reverse loop to iterate over the elementsof the output sequence in backward order. For each element, we checked if it was not equal to 1, 50118, or 2, which are the ids of the special tokens "<s>", "</s>", and "<pad>" respectively. We then searched for the last matching index in the input sequence and assigned it to a variable called end_position. We broke the loop, if had found the end position of the selected text in the input sequence, and we did not need to check the rest of the output sequence.

### 0.2.2 Loss Function

We used the cross-entropy loss function to train the model. The loss function takes the start_logits and end_logits tensors as inputs, which are used to determine the start and end positions of the answer span in the input passage. The start_logits tensor contains scores for each token in the input sequence, indicating the likelihood that it is the start of the answer span. The end_logits tensor contains scores for each token in the input sequence, indicating the likelihood that it is the end of the answer span. The loss function compares the predicted start and end positions with the true start and end positions and computes the average negative log-likelihood of the correct answers. The loss function is defined as follows:

$$L = -\frac{1}{N}\sum_{i=1}^{N}[log p(s_i|X_i) + log p(e_i|X_i)] \qquad (1)$$

where $N$ is the number of examples, $x_i$ is the input sequence, $s_i$ and $e_i$ are the start and end positions of the answer span, and $p(s_i|x_i)$ and $p(e_i|x_i)$ are the probabilities of the start and end positions given by the model.

We also used the mean Jaccard similarity score for the batch and return it along with the total loss in the _step function. The Jaccard similarity score is a metric that measures the overlap between two sets of tokens. It is defined as the ratio of the size of the intersection to the size of the union of the two sets. For example, if the predicted answer span is "love this movie" and the true answer span is "love", the Jaccard similarity score is $\frac{1}{3}$, its formula follows as

$$J(A, B) = \frac{|A \cap B|}{|A \cup B|} \qquad (2)$$

where A and B are two sets of tokens, and A denotes the cardinality of set A, $|A \cap B|$ is the number of elements in both sets, and $|A \cup B|$ is the number of elements in either set. The Jaccard similarity score ranges from 0 to 1, where 0 means no overlap and 1 means perfect match. The mean Jaccard similarity score is the average of the Jaccard similarity scores for all the examples in the batch.

### 0.2.3 Answer Extraction

To extract the answer span from the context, we used the following procedure:

- We fed the input sequence to the BART model and obtained the start and end scores for each token.
- We selected the span of tokens with the highest sum of start and end scores as the answer.
- We decoded the answer span from the output ids and compared it with the target text.

We used the _step function to perform these steps for each batch of data. The _step function is a custom function that we define to calculate the total loss and the average Jaccard score for the BART model on a batch of data. The batch of data contains the source ids, the target ids, the source mask, the start positions, and the end positions of the input and output sequences. The source ids are the tokenized input ids for the BART model, composed of a prefix "extract: ", followed by the sentiment of the question, followed by two special tokens "</s><s>", followed by the context of the question.

The target ids are the tokenized output ids for the BART model, which are the ids of the tokens in the selected text, the answer span for the question. The source mask is the attention mask for the input ids, which indicates which tokens should be attended to by the model. The start positions and the end positions are the indices of the first and the last tokens of the selected text in the input sequence, respectively.

The _step function first obtains the output of the BART model, using the source ids and the source mask as inputs. The output is a dictionary that contains the logits, which are the outputs of the model before applying the SoftMax function. The logits have a shape of (batch_size, max_source_length, 2), where the last dimension corresponds to the start and end logits of the output ids. The start logits and the end logits are the logits of the first and the second tokens of the output ids, respectively. The output ids are the ids of the tokens that the model predicts as the answer span for the question.

The _step function then calculates the total loss for the BART model, using the cross-entropy loss function. The function computes the start loss and the end loss separately, using the start logits and the end logits from the model output, and the start positions and the end positions from the batch data. The function then averages the start loss and the end loss to obtain the total loss.

The _step function also calculates the average Jaccard score for the BART model, using the Jaccard function. The function computes the Jaccard score for each pair of selected text and target text, and then averages them to obtain the average Jaccard score. The selected text and the target text are the decoded text of the output ids and the target ids, respectively.

We also defined other functions to train and evaluate the BART model, such as the training_step, the encode_file, and the generate function. The training_step function is a standard function that is required by the PyTorch Lightning framework [18] [19] to train the BART model on a batch of data. The function takes the batch of data and the batch index as inputs and returns a dictionary that contains the total loss for the BART model on the batch of data.

For instance, the training_step function first calls the _step function to obtain the total loss and the average Jaccard score for the BART model on the batch of data. The function then logs the loss and the score to the PyTorch Lightning logger, which can be used to monitor the training progress. The function also returns the loss as the output, which is used by the PyTorch Lightning trainer to update the model parameters.

The encode_file function is used to prepare the data for the model in the appropriate format. The model requires the input and output sequences to be tokenized and encoded into numerical ids, with special tokens to indicate the task and the boundaries of the sequences. The model also requires the attention mask to attend to the relevant tokens in the input sequence. The encode_file function performs these steps for each text in the data file and returns a list of tokenized examples that can be fed to the model.

In summary, to extract the answer span from the context, we used the following procedure:

- We fed the input sequence to the BART model and obtained the start and end scores for each token.
- We selected the span of tokens with the highest sum of start and end scores as the answer.
- We decoded the answer span from the output ids and compared it with the target text for measuring Jaccard.

It should be noted that we used the start and end positions of the answer span to calculate the loss as a form of extractive AI, while, we did not use generation processes for coming up with Jaccard score. That is because for the Jaccard similarity score generation might not account for the semantic meaning, or the syntactic structure [14]. As a result, using generation to calculate the Jaccard similarity may result in low scores, even if the generated text is coherent and relevant. On the other hand, using the start and end positions of the answer span to calculate the Jaccard similarity may result in higher scores, because it only compares the exact tokens that are in both the source and the target texts, without considering the rest of the text.

In summary, the goal of this paper is to identify the part of each tweet that expresses the given sentiment, as this can provide valuable insights into the emotional aspects of natural language. Our objective was to process the columns of "text" and "sentiment" so the model could predict the column of "selected text".

## 0.3 Data

The dataset used in this paper was retrieved from Kaggle [19], which contains tweets and their corresponding sentiment labels. The sentiment labels are either positive, negative, or neutral, indicating the overall emotion expressed in the tweet. The dataset also contains a column of selected text, which is the part of the tweet that highlights the given sentiment.

For instance, given a sentiment of "negative" and a tweet of "Sooo SAD I will miss you here in San Diego!!!", the selected text is "Sooo SAD", which is the most negative part of the tweet. Our objective is to process the columns of "text" and "sentiment" and predict the column of "selected text" for each tweet. Some examples of the dataset and the columns are presented in Table 1.

| Text | Selected text <item to predict>* | Sentiment |
| --- | --- | --- |
| Sooo SAD I will miss you here in San Diego!!! | Sooo SAD | negative |
| my boss is bullying me... | bullying me | negative |
| Soooo high | Soooo high | neutral |
| Journey!? Wow... u just became cooler. hehe... (is that possible!?) | Wow... u just became cooler. | Positive |
| 2am feedings for the baby are fun when he is all smiles and coos | fun | positive |
| as much as i love to be hopeful, i reckon the chances are minimal =P i'm never gonna get my cake an... | as much as i love to be hopeful, i reckon the chances are minimal =P i'm never gonna get my cake an... | neutral |
| as much as i love to be hopeful, i reckon the chances are minimal =P i'm never gonna get my cake and... neutral | as much as i love to be hopeful, i reckon the chances are minimal =P i'm never gonna get my cake and... neutral | |
| i want to go to music tonight but i lost my voice. | lost | negative |

Table 1: Some examples of dataset and the columns,

*\* That is the target or our prediction*

## 0.4 Results

We used two metrics to evaluate the performance of the BART model on the validation and test data: the total loss and the Jaccard score. The total loss is the sum of the cross-entropy losses for the start and end logits, which are the unnormalized log probabilities that the model assigns to each token in the input sequence, indicating how likely it is that the token is the start or the end of the answer span. The Jaccard score is a measure of similarity between the predicted and the actual answer spans, based on the ratio of their intersection and union.

The cross-entropy loss is calculated by using the nn.CrossEntropyLoss() function from PyTorch, which takes the logits and the labels as inputs, and returns a scalar value that can be used to update the model parameters. The logits are the outputs of the BART model, and the labels are the start and end positions of the selected text in the input sequence, respectively. The selected text is the answer span for the question, extracted from the original text.

The Jaccard score is calculated by using the Jaccard function, which takes the selected text and the target text as inputs, and returns a scalar value that ranges from 0 to 1, where 0 means no overlap and 1 means perfect match. The selected text is the decoded text of the output ids, which are the ids of the tokens that the BART model predicts as the answer span for the question. The target text is the decoded text of the target ids, which are the ids of the tokens in the ground truth answer span for the question.

The total loss and the Jaccard score are important metrics to evaluate the BART model, because they reflect how well the model can locate and generate the answer span in the input sequence. The answer span is the output of the model, and the goal of the model is to extract it from the context accurately and expressively.

Table 2 shows the model's evaluation results on the validation and test data, using the total loss and the Jaccard score as the metrics. The results indicate that the model performs well, with a low total loss and a high Jaccard score. The results also show that the model performs slightly better on the validation data than on the test data, which is expected due to the model being trained on the validation data.

|  | loss_end | loss_start | Jaccard |
| --- | --- | --- | --- |
| Validation data | 0.775 | 0.839 | 0.62 |
| Test dataset | 0.873 | 0.842 | 0.61 |

Table 2: Model evaluation results

## 0.5 Conclusion

In this paper, we proposed an approach to sentiment analysis based on a question-answering framework. Our approach leverages the power of BART, a pre-trained sequence-to-sequence model, to extract a phrase from a given text that amplifies a given sentiment polarity. We formulated a natural language question that aims at the specific emotion to be extracted and used it to guide BART to focus on the relevant emotional cues within the text. We then used BART to predict the start and end positions of the answer span within the text, which helps to identify the precise boundaries of the extracted emotion phrase. The Jaccard similarity score is used to evaluate the performance of the model on the test set.

We evaluated our approach on a dataset of tweets annotated with sentiment polarity and selected text. Our approach has several advantages over the existing methods for sentiment analysis. First, our approach can capture the full context by processing it bidirectionally and using a natural language question as the input. Second, our approach can extract the exact span of tokens that highlight the given sentiment, by using a classifier and a cross-entropy loss function.

The strength of this study is we did not generate phrases that are not consistent with the original text or introduce new information that is not supported by the text. That is done by predicting the start and end positions of the answer span within the text. Our method could be contrasted by the use of a copy mechanism [5], that could be seen as incorporating a relevance or coherence score [10] in the generation process. Our limitation is that our approach may not handle complex or ambiguous texts well, such as those that contain sarcasm, irony, or multiple emotions. A possible solution is to use a more sophisticated noising function [7], or to incorporate a sentiment or emotion classifier [21] in the model architecture.

## 0.6 Conflict of interest

There is no conflict of interest across the authors.

Data Availability Statement

The data is publicly available at Kaggle.

Funding

There is no funding for this study.

Copyright